\documentclass[lettersize,journal]{IEEEtran}
\usepackage{amsmath,amsfonts}
\usepackage{algorithmic}
\usepackage{array}
\usepackage{textcomp}
\usepackage{stfloats}
\usepackage{url}
\usepackage{verbatim}
\usepackage{graphicx}
\def\BibTeX{{\rm B\kern-.05em{\sc i\kern-.025em b}\kern-.08em
    T\kern-.1667em\lower.7ex\hbox{E}\kern-.125emX}}
\usepackage{balance}

\usepackage{marvosym}
\usepackage{indentfirst}
\usepackage{multirow}
\usepackage{rotating}
\usepackage{booktabs}
\usepackage{float}
\usepackage[caption=false,font=normalsize,labelfont=sf,textfont=sf]{subfig}
\usepackage{amssymb}
\usepackage{amsmath}
\usepackage{xcolor}
\usepackage{enumitem}
\usepackage{makecell}
\usepackage{threeparttable}
\usepackage{diagbox}
\usepackage{algorithm}
\usepackage{algorithmic}

\usepackage{hyperref}
\hypersetup{
  colorlinks,
  citecolor=blue,
  linkcolor=red,
  urlcolor=blue
}

\definecolor{myyellow}{rgb}{1, 0.8745, 0.2588}

\begin{document}
\title{Volumetric Medical Image Segmentation via Scribble Annotations and Shape Priors}

\author{\IEEEauthorblockN{
Qiuhui Chen\IEEEauthorrefmark{2}, 
Haiying Lyu\IEEEauthorrefmark{3},
Xinyue Hu\IEEEauthorrefmark{4}, 
Yong Lu\IEEEauthorrefmark{3}, and
Yi Hong\thanks{*Yi Hong is the corresponding author. \href{mailto:yi.hong@sjtu.edu.cn}{yi.hong@sjtu.edu.cn}}
\IEEEauthorrefmark{2}\IEEEauthorrefmark{1}
}

\IEEEauthorblockA{\IEEEauthorrefmark{2}Department of Computer Science and Engineering, Shanghai Jiao Tong University, Shanghai, China \\
\IEEEauthorrefmark{3} Department of Radiology, Ruijin Hospital, Shanghai Jiao Tong University of Medicine, Shanghai, China \\
\IEEEauthorrefmark{4}School of Biomedical Engineering, Shanghai Jiao Tong University, Shanghai, China
}
}

% \author{Qiuhui Chen, Xinyue Hu and Yi Hong}

% \author{Qiuhui Chen\inst{1} \and 
% Yi Hong\inst{1}\textsuperscript{(\Letter)}
% }

% \institute{
% $^1$Department of Computer Science and Engineering, Shanghai Jiao Tong University 
% \email{yi.hong@sjtu.edu.cn}\\
% }

% \markboth{Journal of \LaTeX\ Class Files,~Vol.~18, No.~9, September~2020}%
% {How to Use the IEEEtran \LaTeX \ Templates}

\maketitle

\begin{abstract}
Recently, weakly-supervised image segmentation using weak annotations like scribbles has gained great attention in computer vision and medical image analysis, since such annotations are much easier to obtain compared to time-consuming and labor-intensive labeling at the pixel/voxel level. However, due to a lack of structure supervision on regions of interest (ROIs), existing scribble-based methods suffer from poor boundary localization. Furthermore, most current methods are designed for 2D image segmentation, which do not fully leverage the volumetric information if directly applied to each image slice. In this paper, we propose a scribble-based volumetric image segmentation, Scribble2D5, which tackles 3D anisotropic image segmentation and aims to its improve boundary prediction. To achieve this, we augment a 2.5D attention UNet with a proposed label propagation module to extend semantic information from scribbles and use a combination of static and active boundary prediction to learn ROI's boundary and regularize its shape. Also, we propose an optional add-on component, which incorporates the shape prior information from unpaired segmentation masks to further improve model accuracy. Extensive experiments on three public datasets and one private dataset demonstrate our Scribble2D5 achieves state-of-the-art performance on volumetric image segmentation using scribbles and shape prior if available. Our code is available online:
\href{https://github.com/Qybc/Scribble2D5}{https://github.com/Qybc/Scribble2D5}
\end{abstract}

\begin{IEEEkeywords}
Weakly-supervised Learning, Scribble Annotation, Volumetric Image Segmentation, Shape Prior.
\end{IEEEkeywords}

\section{Introduction}

\begin{figure}[t]
\includegraphics[width=0.48\textwidth]{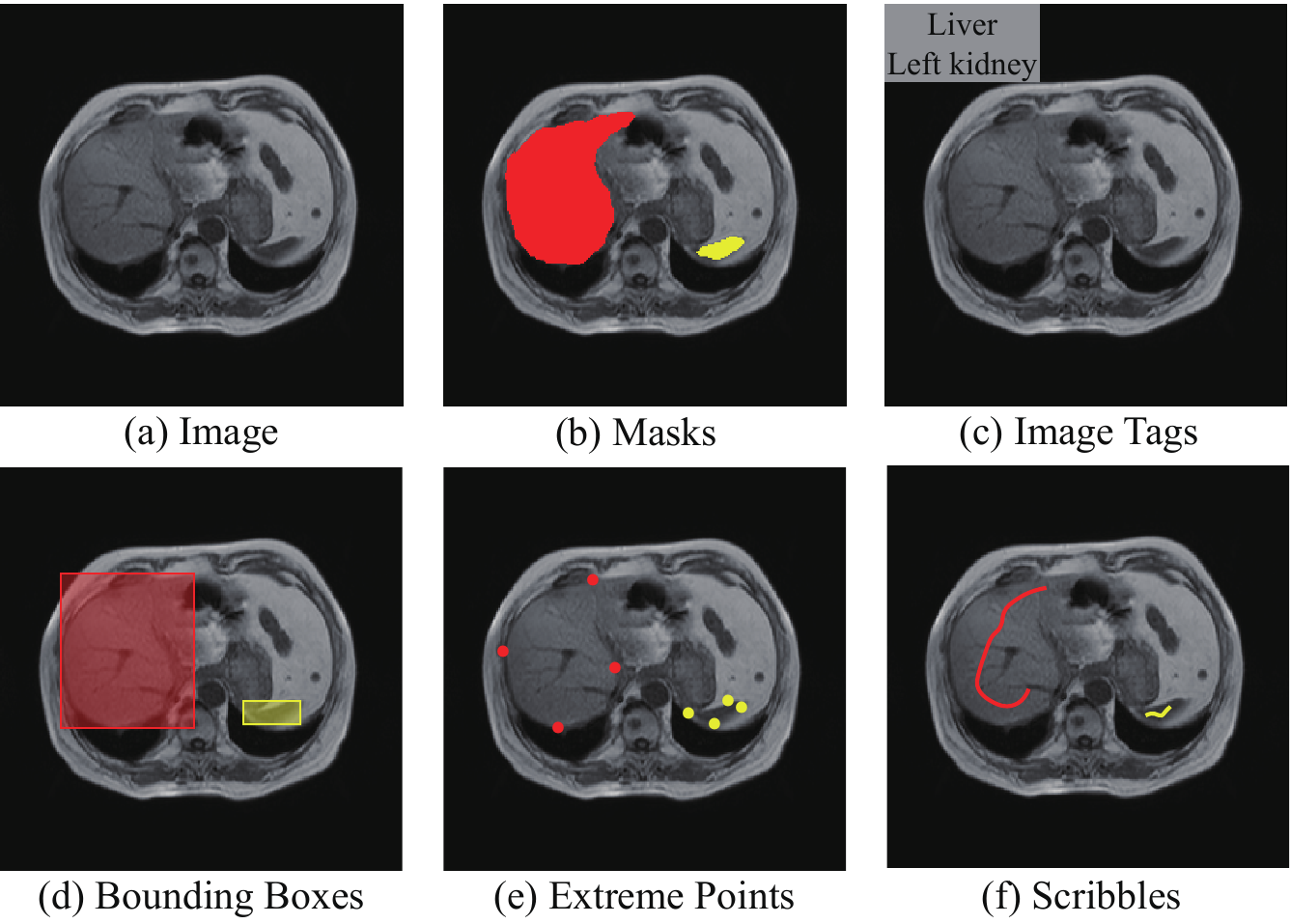}
\caption{Examples of different types of annotations used in medical image segmentation. The axial image slice is sampled from the Combined Healthy Abdominal Organ Segmentation (CHAOS) dataset~\cite{luo2022scribble}. ({\textcolor{red}{Red}}: liver, {\textcolor{myyellow}{yellow}}: left kidney) 
% \fixme{Masks, bounding boxes}
} 
\label{fig:weak}
\end{figure}

Deep-learning-based segmentation networks have achieved impressive accuracy in many medical applications, especially in a fully-supervised manner~\cite{shapey2019artificial,zhou2018unet++}. However, to train a deep segmentation network, such methods often require a large number of dense annotations at pixel or voxel levels, as the masks shown in Fig.~\ref{fig:weak}(b). In practice, dense manual annotations for medical images are difficult to obtain because annotating at image pixels or voxels is time-consuming and needs medical expertise to provide high-quality segmentation masks. Another choice is using fully-unsupervised segmentation methods~\cite{xia2017w,dey2021asc}, which have shown promising segmentation results. However, their performance gap with respect to fully-supervised approaches is too large to make them practical. Therefore, weakly-supervised approaches by using weak annotations have gained great attention, which can greatly reduce the workload of manual annotations and produce promising results that are comparable to fully-supervised segmentation approaches.

Figure~\ref{fig:weak} presents several commonly-used weak annotations, including image-level annotations~\cite{xu2015learning,ahn2018learning}, bounding boxes~\cite{rajchl2016deepcut,shapey2019artificial}, extreme points~\cite{maninis2018deep,roth2021going}, and scribbles~\cite{lin2016scribblesup,tang2018normalized,dorent2020scribble,valvano2021learning}. Compared to image-level and bounding box annotations, scribbles provide rough positions of Regions of Interest (ROIs) to allow for a better location. Also, annotating by scribbles is more flexible than using bounding boxes and extreme points, especially for ROIs with irregular shapes. Annotators have no need of knowing the exact boundaries of ROIs, which benefits users since locating ROIs' boundaries is not an easy task and requires more expertise. With basic training, users with no medical background can quickly learn how to make scribble annotations, making this type of annotation useful in practice. Therefore, we choose scribbles as our weak annotations. 
%In addition, extreme point annotations are more suitable for convex shapes and may not work for non-convex ones. 

Although using scribble annotations for medical image segmentation is beneficial in many aspects, there are several challenges faced by scribble-based learning methods. Firstly,  
scribbles are often sparse with no structure information of ROIs; as a result, scribble-based methods have difficulty in accurately locating the ROI's boundaries~\cite{tang2018normalized}.  
Moreover, existing scribble-based methods are typically designed for 2D images~\cite{lin2016scribblesup,wang2019boundary,zhang2020weakly,valvano2021learning}, which do not fully leverage the whole image volume by directly applying on image 3D volumes, with missing connections between slices. Preliminary work in~\cite{dorent2020scribble} performs 3D segmentation by using transfer learning, which alternatively learns by using mask annotations in the source domain and scribbles in the target domain. Also, in practice, many clinical problems collect anisotropic medical image volumes, with a much larger voxel spacing in one view than others. We aim to tackle these problems and build a weakly-supervised segmentation network, which suits anisotropic medical volumes with improved boundary localization, and operates automatically at the inference stage, with no need of providing any scribble inputs.

To achieve this goal, we propose a volumetric segmentation network called Scribble2D5. This model adopts a 2.5D attention UNet~\cite{shapey2019artificial} to handle anisotropic medical volumes with different voxel spacings. To amplify the influence of the sparse scribbles in volumetric segmentation, we use a label propagation module based on supervoxels to generate 3D pseudo masks from scribbles for supervision. To address the boundary localization issue, we propose using the combination of learning both static and active boundaries via predicting edges in 3D and a proposed active boundary loss in 3D based on active contour model~\cite{chen2019learning}. Also, we consider shape priors via 
shape descriptors~\cite{fang20153d} and skeleton context~\cite{jiang2015informative}
to further improve the quality of the boundary localization. This add-on component fully leverages existing unpaired segmentation masks while incorporating expert knowledge without requiring additional annotations.

%In this paper, we introduce a novel training strategy in the context of weakly supervised learning for medical volumetric segmentation as an extension of. We train a model for medical semantic segmentation using scribbles, 

%In addition, compared to~\cite{chen2022scribble2d5}, we consider the case of missing scribble annotations on some slices or reducing the manual work via an adaptive annotation with the watershed techinque. Finally, the method was validated using private datasets, achieving excellent accuracy scores comparable to repeated annotations performed by clinicians. 

% We evaluate our methods on three datasets, including ACDC dataset~\cite{bernard2018deep} for cardiac segmentation, VS dataset~\cite{shapey2021segmentation} for tumor segmentation, CHAOS dataset~\cite{kavur2021chaos} for abdominal organ segmentation, and \fixme{a private dataset for ...}. For both ACDC and CHAOS datasets, our method outperforms the current state-of-the-art (SOTA) by large margins on three different evaluation metrics; and on the VS dataset, our method achieves better performance in Dice compared to SOTA. \fixme{For the private dataset .... } 
% Our method reduces the performance gap between weakly-supervised and fully-supervised approaches, and its application to our private dataset demonstrates the potential use in practice. 

This paper is an extension of our conference paper~\cite{chen2022scribble2d5}, by adding an optional shape prior component and performing extensive experiments, including evaluation on an additional private dataset, comparison with more baselines, study on how to handle missing or partial scribble annotations, and the comparison between real and generated scribbles. Overall, the contributions of this paper are summarized as follows:
\begin{itemize}[noitemsep,nolistsep]
    \item We propose a scribble-based volumetric image segmentation network, Scribble2D5, which handles anisotropic medical scans and improves boundary localization via a 3D label propagation, static and active boundary prediction and regularization, and shape priors learned from unpaired segmentation masks.
    \item We achieve SOTA performance compared to nine baseline models on three public datasets (i.e., ACDC~\cite{bernard2018deep} for cardiac segmentation, VS~\cite{shapey2021segmentation} for vestibular schwannoma tumor segmentation, CHAOS~\cite{kavur2021chaos} for abdominal organ segmentation) and one private dataset for the segmentation of pituitary with tumor.
    \item We conduct comprehensive experiments to evaluate the performance of our method, not only using multiple datasets, including both public and private datasets to demonstrate its practicality, but also studying the effect of using partial, real, or generated scribbles. 
   % We propose a 3D label propagation method for 3D pseudo mask generation.  
   % \item We propose a shape-guided entropy minimization objective, which can be readily used for integrating shape priors to any segmentation network, as long as we have the same objects of medical segmentation datasets.
  
    %\item We investigate diverse learning scenarios, such as: learning from different extents of weak annotations (i.e., weakly supervised learning); and learning also with strong supervision of unpaired segmentation masks (i.e., multi-task learning).
\end{itemize}

\begin{figure*}[t]
\includegraphics[width=\textwidth]{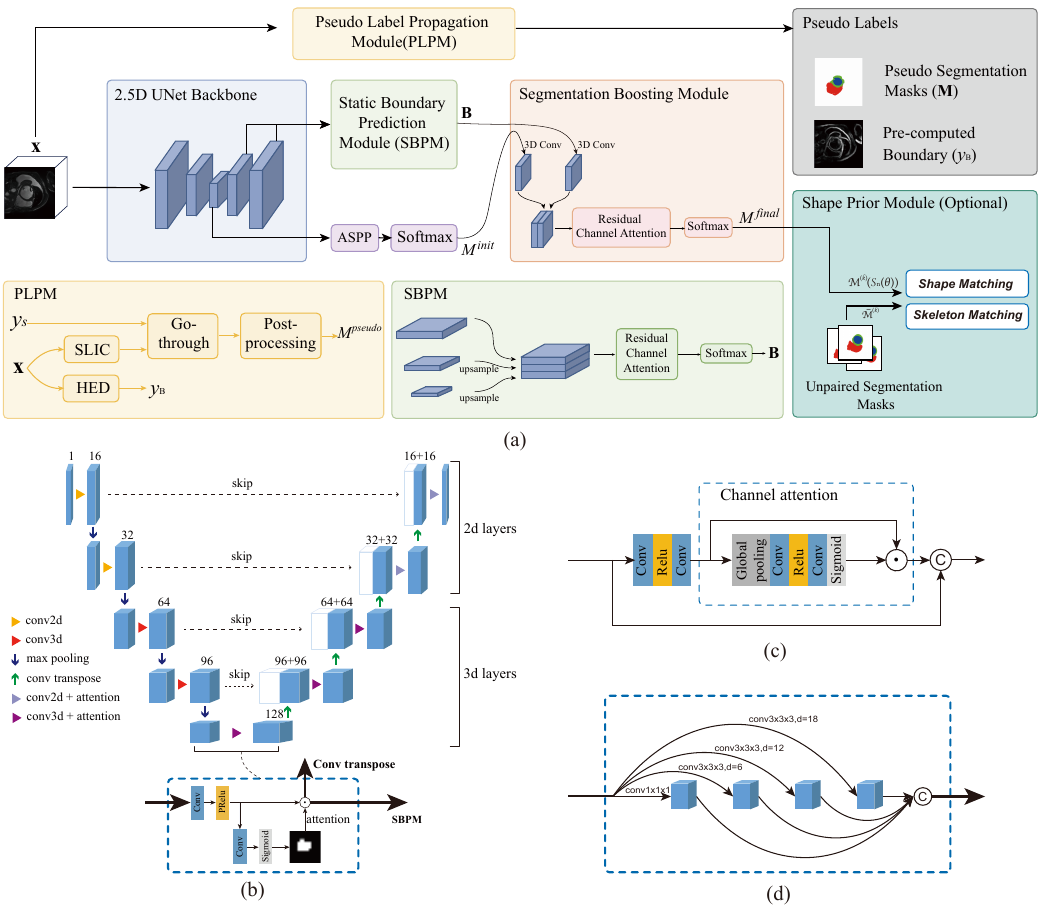}
\caption{Overview of our Scribble2D5 model, including five components: 1) pseudo label propagation module (PLPM, yellow block), which generates image boundaries and pseudo-3D segmentation masks based on scribble annotations; 2) 2.5D attention U-Net as our backbone network (blue block, see details in (b)); 3) static boundary prediction model (SBPM, green block), which uses boundary information $y_B$ pre-computed by PLPM for supervision; 4) segmentation boosting module (SBM, orange block), which further considers active boundaries via an active boundary loss; 5) shape prior model (cyan block), which regularizes segmentation masks with shape prior. Both SBPM and SBM modules use residual channel attention blocks as shown in (c). (d) ASPP (Atrous Spatial Pyramid Pooling) block used between U-Net and SBM. (Best viewed in color)
% \fixme{the active contour loss is missing} Loss Function都没体现在图上
} 
\label{fig:overview}
\end{figure*}

\section{Related Work}

%A lot of research works have been focusing on developing learning algorithms that rely less on high-quality annotations~\cite{cheplygina2019not, tajbakhsh2020embracing}. Below, we briefly review recent weakly supervised methods that use scribbles to learn image segmentation. Then, we discuss what are the advantages of our method compared to other scribble-based methods. Finally, we discuss the difference between the shape priors that are an integral part of our segmentor and other shape prior supervision modules.

In this section, we briefly review recent works on weakly-supervised image segmentation using scribble annotations. Also, we discuss image segmentation with shape priors, which is a useful add-on component to existing methods. 

\subsection{Learning from scribble supervision}

Scribbles are sparse annotations that have been successfully used in semantic segmentation. The segmentation accuracy of scribble-based methods is approaching full-supervised methods in both computer vision and medical image applications~\cite{valvano2021learning,luo2022scribble,zhang2022cyclemix}. However, scribbles lack structure and shape information of objects or ROIs, which makes the accurate segmentation of object boundaries a challenging task for existing methods~\cite{lin2016scribblesup}. To address this problem, propagating scribble annotations to generate masks for full supervision is a commonly-used strategy. In~\cite{lin2016scribblesup,ji2019scribble}, scribble annotations are expanded to adjacent pixels with similar intensity using graph-based methods. In~\cite{can2018learning}, a two-step procedure is used to first estimate the labels for unannotated pixels of ROIs based on scribbles and then refine the predictions by using Conditional Random Fields (CRF). The main limitation of these approaches is the inaccurate relabeling step, which is time-consuming and brings labeling errors for supervising the learning of following-up segmentation models. Thus, other researchers have investigated alternatives to avoid this relabeling step, such as using a CRF-based loss regularizer~\cite{tang2018regularized}, a post-processing step with CRF~\cite{chen2017deeplab}, or a trainable CRF layer~\cite{zheng2015conditional}. 

Our method avoids the data relabeling step by directly learning a mapping from images to segmentation masks, without using the expensive CRF-based post-processing. We cope with unlabelled regions of the image with the help of a label propagation module based on supervoxels and 3D image edges. 
%Differently from the work in~\cite{tang2018regularized}, our label propagation module does not rely on CRF and \fixme{allows handling both long-range and short-range inconsistencies in the predicted masks.why?}
Concurrent to our work on weakly-supervised 3D medical image segmentation, Kervadec et al.~\cite{kervadec2019constrained} propose an unsupervised regularization term of the loss function to constrain the 3D volume size of the target region. Luo et al.~\cite{luo2022scribble} propose a dual-branch network to dynamically mix-up pseudo-labels by mixing the two branches’ outputs and use the generated pseudo labels to supervise the network training. Zhang and Zhuang~\cite{zhang2022cyclemix} propose a mixup augmentation of image and scribble supervision and a regularization term of supervision via cycle consistency. These methods mainly work on 2D slices when handling 3D images. Although the work in~\cite{kervadec2019constrained} regularizes the volume size of the segmentation output, its network takes 2D slices as inputs. Differently, our scribble2D5 tackles 3D anisotropic images as inputs, considering 3D shapes of ROIs to treat objects as a whole for learning.

Recently, the Segment Anything Model (SAM) proposed in~\cite{kirillov2023segment} has achieved great success in segmenting natural images in computer vision. A couple of following works~\cite{he2023accuracy,wu2023medical} study its application or extension to medical images, which is still at an early stage and needs more effort to work well in the medical domain. According to our experience working with a private dataset, our scribble2D5 is easy to use in practice and has the flexibility of being adopted by different medical image segmentation tasks.  

\subsection{Shape Priors in Deep Medical Image Segmentation}

In semantic segmentation, incorporating shape prior knowledge into pixel-level segmentation is an efficient way to address object occlusion or low image quality issues. 

%there has been considerable interest in incorporating prior knowledge about organ shapes to obtain more accurate and plausible results. 

A common way to incorporate shape priors into image segmentation is matching the predicted masks with those provided in the shape priors, by using an additional module, such as the multi-scale attention gates used in adversarial training~\cite{zhang2020accl}, a PatchGAN discriminator~\cite{isola2017image}, the persistent homology~\cite{clough2019topological}, etc. Others~\cite{oktay2017anatomically} demonstrate that a data-driven shape prior can be learned through a convolutional autoencoder from unpaired segmentation masks and used as a regulariser to train a segmentation network. Similarly, a variational autoencoder (VAE)~\cite{kingma2013auto} is adopted to learn shape priors~\cite{dalca2018anatomical}, which has partial weights shared with a segmentation model. Other approaches consider shape priors 
%in the post-processing step, 
in the training with a regulariser~\cite{yue2019cardiac} or a differentiable penalty~\cite{kervadec2019constrained}, or at the inference stage via adjustment using VAEs~\cite{painchaud2019cardiac} or denoising autoencoders~\cite{larrazabal2020post}. 

Considering the stability issue of adversarial learning and the variation of masks for the same type of objects, we turn to the traditional shape descriptors~\cite{celebi2005comparative}, which are more robust and invariant across image modalities or subject populations. These shape descriptors are integrated into our main segmentation network as an optional component.

\section{Methodology}

Figure~\ref{fig:overview} presents the framework of our proposed Scribble2D5, a weakly-supervised image volume segmentation network based on scribble annotations and shape priors. Scribble2D5 uses a 2.5D attention UNet~\cite{shapey2019artificial} as the backbone network, which is augmented by four modules, i.e., 1) a Pseudo Label Propagation Module (PLPM) for generating 3D pseudo masks and boundaries for supervision, 2) a Static Boundary Prediction Module (SBPM) for incorporating object boundary information from images, 3) a Segmentation Boosting Module (SBM) for further considering active boundaries via an active boundary loss, and 4) an optional Shape Prior Module (SPM) for incorporating shape prior knowledge and encouraging the final prediction to be more accurate and realistic. 

\subsection{Backbone}

The image volumes we study in the experiments are anisotropic with different voxel spacings, which is very common in practice. In our dataset, the in-plane resolution within a slice is about four times the thickness of a slice. Since 2D CNNs ignore the important correlations between slices and 3D CNNs typically handle isotropic image volumes, we choose a 2.5D neural network that considers the anisotropic properties of an image volume. In particular, we adopt an attention UNet2D5~\cite{shapey2019artificial} as our backbone network, which augments UNet2D5 by adding an attention block at each deconvolutional layer, as shown in Fig.~\ref{fig:overview}(b). Specifically, at the top two layers of both encoder and decoder branches, we have 2D convolutional operations; while at other layers, the feature maps are roughly isotropic, which are suitable for 3D convolutions. The attention blocks are noted by purple triangles in Fig.~\ref{fig:overview}(b). Their attention maps are estimated via two layers of convolutions, i.e., one with Peakly ReLU (PReLu) and the other with a Sigmoid activation function. This 2.5D network suits for all images in our experiments. In practice, the number of 2D convolution layers can be adjusted according to image resolution; if the image volume is isotropic with an equal voxel spacing, the 2.5D UNet degenerates to 3D UNet.

\subsection{Pseudo Label Propagation Module (PLPM)}

To augment the supervision effect of weak annotations like scribbles and fully leverage the input image, in this pre-processing step, we generate a 3D Pseudo mask using  scribble propagation and a 3D static boundary label, which will be used later for guiding the learning of our Scribble2D5 model.   

\subsubsection{3D Pseudo Mask Generation}
Scribble annotations are often sparse, which cover only a small amount of pixels on each slice of an image volume. As a result, the supervision information from scribbles is not strong enough to produce satisfied guidance, as reported in $\text{UNet}_\text{PCE}$~\cite{tang2018normalized}. To address this issue, we leverage the technique of supervoxels to magnify the effects of scribble annotations in 3D. In particular, we adopt SLIC~\cite{achanta2012slic}, which generates supervoxels from images by using an adaptive k-means that considers both image intensity and distance similarities when clustering. We then collect those supervoxels where scribbles pass through, resulting in 3D pseudo segmentation masks for our regions of interest (ROIs).
 
When generating the 3D Pseudo mask, we assume the scribble annotations are available on all image slices. However, in practice, annotating scribbles on all slices is still time-consuming and demands lots of labor effort. A possible solution is to select some slides for annotating and use a strategy to expand these annotations to other slides. In particular, we assume the slide centered in the region of interest (ROI) contains the most information compared with other slides; therefore, we choose it as our starting point for annotation and label it "annotated". Then, we gradually divide the remaining slides into two groups, i.e., the annotated group and the un-annotated group. Each time we firstly compute the Structural Similarity Index Measure (SSIM) between these two groups and from the un-annotated group we select the one with the highest SSIM score into the annotated group. This process continues until the number of slides in the annotated group reaches to its maximum value. 

The next step is to propagate the scribbles on the annotated slides to other un-annotated ones. One choice is using a 3D anisotropic watershed approach~\cite{vincent1991watersheds}, which considers the different voxel spacing when flooding to other slides. Then an erosion is adopted to reduce the width of generated annotations, which makes them more like scribbles and reduces false positives of generated annotations. Another choice is using random walk~\cite{spitzer2001principles} based on an anisotropic diffusion. This method is slower then the watershed method; however, it produces better results as shown in our experiments. In this way, we can handle the case of missing scribble annotations on some slides and provide an approach to reduce the manual work of making annotations when preparing the training set. After having scribbles on all slides of an image, we can generate its 3D pseudo mask as discussed before.

\subsubsection{3D Static Boundary Label Generation}
Except for the pseudo mask we generate from the scribble annotations, we also generate the pseudo static boundary of ROI from an image volume by stacking 2D edges detected on each slice. This boundary is static since it is pre-computed from the image and keeps unchanged during training, which is different from the active boundary we will discuss later. To obtain 2D edges, we directly use an existing method, i.e., HED~\cite{xie2015holistically}, which is pre-trained on the generic edges of BSDS500~\cite{arbelaez2010contour}. 

As a result, this PLPM component generates 3D pseudo masks from scribbles for ROI segmentation and pre-computed boundaries for static boundary prediction, respectively.

\subsection{Static Boundary Prediction Module (SBPM)}

This module encourages the backbone network to extract image features with rich boundary structures at different scales. Following~\cite{zhang2020weakly}, we collect feature maps from different layers of the network decoder, and concatenate these 2D and 3D features at different resolutions right after one convolutional layer with a filter of size $1 \times 1 \times 1$. To fuse these features, we feed them to a residual channel attention block (as shown by a green square in Fig.~\ref{fig:overview}(a)) and a $1 \times 1 \times 1$ convolutional layer to produce a boundary map $b$ in 3D. Under the supervision of the previously generated 3D pseudo boundary $y_B$, the network is trained with a binary cross entropy loss on the network output $B$: 
% BCE Loss
\begin{equation}
\mathcal{L}_{bry}(y_B, B)= -(y_B \textit{log}B+(1 - y_B) \textit{log}(1 - B)).
\label{eq:bry-loss}
\end{equation}
% where $N$ is the number of classes for segmenting ROIs. 
This SBPM module only generates boundaries of images to supervise the learning of our backbone network. To obtain the masks of ROIs, we need the following boosting module. 
% \fixme{how to know $B_c$ and $b_c$?} 

\subsection{Segmentation Boosting Module (SBM)}
This module performs segmentation under the supervision of the previously generated pseudo mask with supervoxels and a regularization on segmentation output. The segmentation includes two stages, i.e., an initial segmentation and a final one with further considering both static and active boundaries. 

To predict a preliminary mask, we employ a dense atrous spatial pyramid pooling (DenseASPP, as shown in Fig.~\ref{fig:overview}(d)) block~\cite{yang2018denseaspp}, right after the bottom layer of the backbone network, which enlarges its receptive fields by utilizing different dilation rates, as shown in Fig.~\ref{fig:overview}(a). In this block, the convolutional layers are connected in a dense way to cover a larger scale range without significantly increase the model size. Then we adopt two additional 3D convolutional layers followed by a $1 \times 1 \times 1$ convolution, resulting in the initial prediction $M^{init}$, which is supervised by the generated pseudo mask $M^{pseudo}$. 
% \fixme{please make sure the notations here consistent with those in Fig.1.} 
Considering the oversegment nature of supervoxels, one supervoxel may be selected by multiple different classes. To avoid this confusion, we only consider those supervoxels with a unique label, which are set to be 1 in the mask $M^{voxel}$ with others being zeros. Therefore, we use the following partial cross entropy to supervise the initial segmentation result:
\begin{equation}
\begin{aligned}
    \mathcal{L}_{seg}(& M^{init}, M^{pseudo}, M^{voxel}) \\
                      & = -\sum_{c=1}^{N} M^{voxel}_c \cdot M^{pseudo}_c \textit{log}(M^{init}_c).
\end{aligned}
\label{eq:seg_init}
\end{equation}
Here, $N$ indicates the number of classes in the segmentation. This loss function allows early feedback to fasten the convergence of our network.

To refine the initial mask prediction and obtain a boundary-preserving mask for a final prediction, we merge outputs from the boundary prediction module with those from the initial mask prediction for refinement. These feature maps are fed to a residual channel attention block, followed by a $1 \times 1 \times 1$ convolutional layer to produce the final mask prediction $M^{final}$. Similarly, we use the partial cross-entropy loss to predict the final mask under the supervision of the generated pseudo mask $M^{pseudo}$.

\noindent
\textbf{Active Boundary (AB) Loss.}
The pseudo masks are imperfect because supervoxels are coarse segmentation masks of ROIs and have oversegment issues, resulting in the potential of having many false positives. To mitigate this issue, we propose regularizing the surface and volume of the 3D segmentation region by extending the 2D active contour loss~\cite{chen2019learning} to a 3D version. We apply an AB loss as follows:
\begin{equation}
\mathcal{L}_{AB} = \textit{Surface} + \lambda_{1} \cdot \textit{Volume}_\textit{In}  + \lambda_{2} \cdot \textit{Volume}_\textit{Out},
\label{eq8}
\end{equation}
where
$\textit{Surface} =\int_{S}|\nabla u| \textit{ds}$ and $u$ is the mask prediction; $\textit{Volume}_\textit{In}=\int_{V}\left(c_{1}-v\right)^{2} \textit{udx}$, $c_1$ is the mean image intensity inside of interested regions $V$, and $v$ is the input image;  $\textit{Volume}_\textit{Out}=\int_{\bar{V}}\left(c_{2}-v\right)^{2} \textit{udx}$ and $c_2$ is the mean image intensity outside of the region. These items are balanced by two hyper-parameters $\lambda_1$ and $\lambda_2$. In the experiments, we set $\lambda_1=1$ and $\lambda_2=0.1$, to emphasize more on the inside region of the volume. This new loss function considers the shape and intensity of an image in 3D, which regularizes ROI's shapes and helps reduce false positives in segmentation.

\subsection{Shape Prior Module (SPM)}
\begin{figure}[t]
\includegraphics[width=0.48\textwidth]{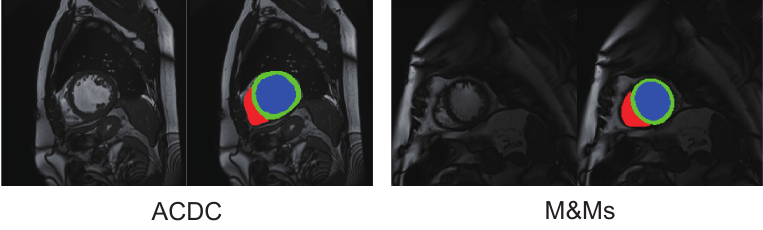}
\caption{Image and mask samples collected from the ACDC and M\&Ms datasets. \textcolor{red}{Red}: left ventricle (LV), \textcolor{green}{green}: myocardium (MYO), \textcolor{blue}{blue}: right ventricle (RV).} 
\label{fig:ACDC_MnMs}
\end{figure}

Another efficient way to mitigate the inaccurate boundary estimation suffered by scribble-based methods is to incorporate shape prior knowledge into the network learning. For instance, from existing public datasets we may obtain some unpaired segmentation masks for ROIs, which can be used to extract shape prior representation for learning.
As shown in Fig~\ref{fig:ACDC_MnMs}, ACDC and M\&Ms are cardiac image segmentation datasets collected from different centers with different MRI scanners, but they are collected to tackle the same segmentation problem, extracting the left (LV) and right ventricles (RV), as well as left ventricular myocardium (MYO) from medical scans. That is, we can fully leverage the masks provided by the M\&Ms dataset to help a better shape extraction for our task on the ACDC dataset. Since the M\&Ms masks are unpaired with the image scans from ACDC, we need a shape descriptor that is invariant across image acquisition centers and scanners. Here, we adopt two types of shape descriptors, i.e., shape moments and skeleton descriptors.

%Obviously, the segmentations have the same distribution. We choose ACDC as the source dataset and the  as the target dataset.

%Our proposed shape prior module build from the hypothesis that certain shape descriptions might be invariant across image acquisition protocols/modalities and subject populations. 

\subsubsection{Shape Moments}

% Shape moments are often used to characterize the shape of an object in computer vision and image processing~\cite{martinez20102d}. Recently, they are demonstrated to be quite useful in some supervised learning tasks~\cite{kervadec2021beyond,bateson2022test}. Before introducing the concept of shape moments, we first restate our problem in mathematical language. 

Given a set of $M$ source images $ I_m:\Omega \in \mathbb{R}^{n_x\times n_y \times n_z}, m=1,2,\cdots,M$, $n_x, n_y, n_z$ are dimensions of an image, we denote their ground truth K-class segmentation for each voxel $i \in \Omega_s$ as a K-simplex vector $\mathbf{y}_{m}(i)=\left(y_{m}^{(1)}(i), \ldots, y_{m}^{(K)}(i)\right) \in\{0,1\}^{K}$. For each voxel $i$, its coordinates in a 3D spatial domain are represented by the tuple $\left(x_{(i)}, y_{(i)}, z_{(i)}\right) \in \mathbb{R}^{3}$. Our goal is to obtain a network $\mathcal{N}_{\boldsymbol{\theta}}: I(i) \mapsto s_{\boldsymbol{\theta}}(i) $ with network parameters $\theta$,  for each voxel $i \in \Omega $, where $s_{\boldsymbol{\theta}}(i)=\left(s_{\boldsymbol{\theta}}^{(1)}(i), \ldots, s_{\boldsymbol{\theta}}^{(K)}(i)\right) \in[0,1]^{K}$, which predicts a softmax probability map for class $k \in {1,2,\cdots,K}$. We define two 3D shape descriptors below to obtain the compact representation of a shape for a given input image $I$ and a specific class $k$.

\noindent
\textbf{Class Ratio $\mathcal{R}$.} This descriptor measures the relative size of a shape.
The ratio of class $k$ can be computed as the percentage of the segmentation volume of this class over the total foreground volume of the input image. To calculate the volume of class $k$, we simply use the summation of its prediction probability, which is a special case of shape moments. As a result, we define the class ratio as 
\begin{equation}
\begin{aligned}
\mathcal{R}^{(k)}(s_{\boldsymbol{\theta}})=\frac{1}{\Omega}\sum_{i \in \Omega} s_{\boldsymbol{\theta}}^{(k)}(i).
\end{aligned}
\end{equation}

% \begin{equation}
% \begin{aligned}
% \mathcal{R}^{(k)}(s_{\boldsymbol{\theta}}):=\frac{1}{\left|\Omega \right|} \mu_{0,0,0}^{(k)}(s_{\boldsymbol{\theta}}) . 
% \end{aligned}
% \end{equation}

\noindent
\textbf{Average Distance to the Centroid $\mathcal{D}$.} This shape descriptor measures on average how far the object spreads around its centroid. Here, we use the standard deviation of pixel coordinates to compute this average distance for class $k$:
\begin{equation}
\begin{aligned}
    & \mathcal{D}^{(k)}\left(s_{\boldsymbol{\theta}}\right)= \\
    & \frac{1}{\Omega}\sum_{i \in \Omega}\left(
    \sqrt[2]{(x_{(i)}^{(k)}-\bar{x}^{(k)})^2}, \sqrt[2]{(y_{(i)}^{(k)}-\bar{y}^{(k)})^2}, 
    \sqrt[2]{(z_{(i)}^{(k)}-\bar{z}^{(k)})^2}
    \right),
\end{aligned}
\end{equation}
where $(\bar{x}^{(k)},\bar{y}^{(k)},\bar{z}^{(k)})$ is the mean coordinate of class $k$.

\begin{figure}[t]
\includegraphics[width=0.48\textwidth]{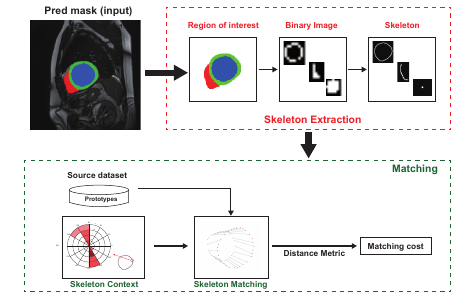}
\caption{The complete scheme for automatic skeleton matching. In the first stage, it extracts regions of interest, which are predicted foreground parts in images. Then in the skeletonization step, it extracts skeleton map of each part, and prunes it using the proposed algorithm. After finding pruned skeleton, it computes skeleton context and finds the nearest match in the database of target prototypes. Finally, it uses mixture discriminant analysis classifier to detect whether it is realistic or not.} 
\label{fig:skeleton_pipeline}
\end{figure}

\subsubsection{Skeleton Descriptors} To make the predicted shape close to the shape described by the unpaired segmentation masks, we propose to extract skeletons from provided and predicted masks and match them for comparison. Overall, our shape matching model includes two steps: the skeleton extraction and skeleton matching, as shown in Fig.~\ref{fig:skeleton_pipeline}. The extraction step takes an image slice and uses a skeletonization strategy to extract the skeleton of the region of interest; later, the matching step measures the distance between the predicted and target masks to identify whether they are similar.

%\fixme{Our proposed scheme, shown in Fig ~\ref{fig:skeleton_pipeline}, consists of two main steps, skeleton extraction and matching. In the first step, we take a slice image and extract its regions of interest. Then using our skeletonization framework, to extract their skeleton. In the second step, using our suggested shape descriptor, skeleton context, to extract features for shape matching part. After matching those parts to the target prototype, based on the distance between them and their matched target prototype, we are able to identify whether they are realistic or not.}

\noindent
\textbf{Skeleton Extraction.}
% Skeleton can provide a good abstraction of a shape, which contains topological structure and its features. Because it is the simplest representation of a shape, there has been an extensive effort among researchers to develop generic algorithms for skeletonization of shapes. However, ~\cite{saha2016survey} claim that since there is no “true” skeleton defined for an object, the literature in skeletonization lack of a robust evaluation. The vast majority of the algorithms are based on Blum’s “Grassfire” analogy and formulation for skeletonization~\cite{blum1973visual}. 
The key point in skeletonization algorithms is to preserve the topology of a shape. We adopt the skeletonizing method proposed in~\cite{rajchl2017employing}, which performs iterative morphological erosion of a segmentation mask to obtain the skeleton of an object. Specifically, for each object in a mask, we iteratively remove the border pixels of an object until a single-pixel edge, line, or point is achieved. Then, we use the gray scale morphological operator to close the generated discontinuous skeleton.

%followed morphology standard skeleton is ation by iterative identification and removal of border pixels, until single-pixel skeleton. 

\begin{figure}[t]
\includegraphics[width=0.48\textwidth]{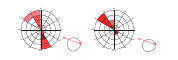}
\caption{Skeleton context of two matched points on different skeletons.} 
\label{fig:skeleton_context}
\end{figure}

\noindent
\textbf{Skeleton Context.}
To describe the extracted skeleton, we use a new descriptor called skeleton context, which is a log-polar histogram formed for each sample point on the skeleton. For each sample point $p_i$, this log-polar histogram treats it as the center, and each bin of the histogram counts the number of sample points at its specific angle and range of distance from the center (i.e. $p_i$). As shown in Fig.~\ref{fig:skeleton_context}, the centers of the red small circles show quite different skeleton context, especially at the right bottom part of the log-polar histogram, where the corresponding segments of two skeletons are different, with missing points on the second skeleton. 

%As shown in Fig.~\ref{fig:skeleton_context}, the skeleton context is a log-polar histogram, formed for each sample point on the skeleton. For each sample point $P_i$, the center of this log-polar histogram is located on that sample point, then each bin in the histogram represents the number of sample points in the specific angle and range of distance from the center (i.e. $P_i$) determined by that bin. We use the notation of $H_{SC} (P_i , r_{k_1}, \theta_{k_2})$, to show the value of skeleton context’s histogram for point $P_i$ , in the $(r_{k_1} , k_2)$ bin. For instance, when $H_{SC} (P_i , r_{k_1}, \theta_{n}) = 10 $, it means that in the distance range of $r_{m-1} \leq r \le r_m$ and in the angle range of $ \theta_{n-1} \leq \theta \le \theta_{n} $ from the point $P_i$ in the skeleton, there are 10 other sample points. Basically, these histograms for each point visualize the distribution of other points on the skeleton with respect to that point, and hence it could play the object descriptor role for the shape matching purposes. As a result, we could use these descriptors as the feature data for matching cost in the next step. The skeleton context can be calculated using the following equation:

\begin{figure}[t]
\includegraphics[width=0.48\textwidth]{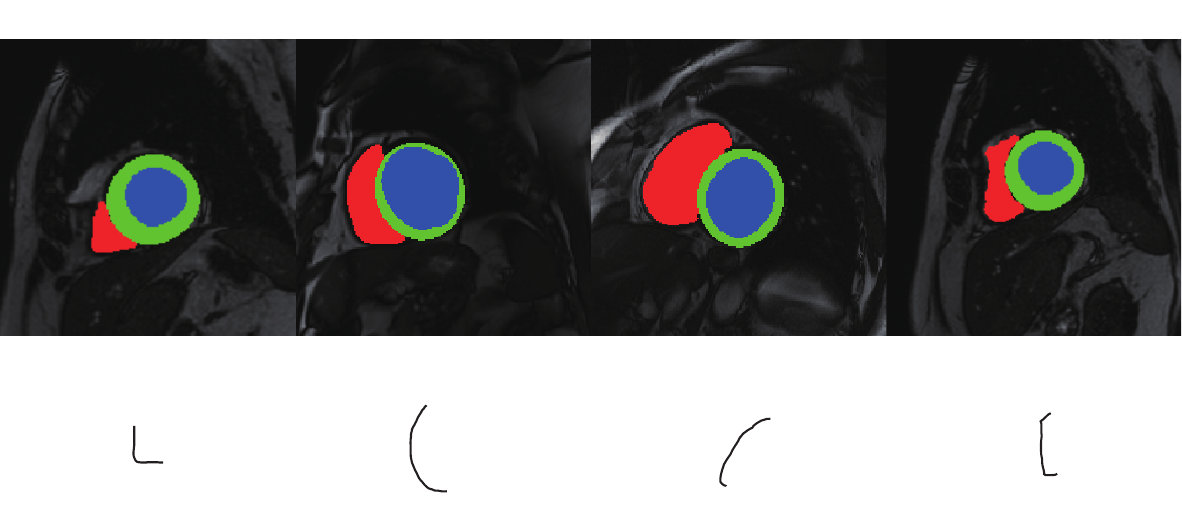}
\caption{Four prototypes (bottom) for the left ventricle (red object), extracted from the masks provided by the M\&M dataset (top).} 
\label{fig:prototypes}
\end{figure}

In particular, we use the notation $H_{SC} (p_i , r_{m}, \theta_{n})$ to present the value of the skeleton context’s histogram centered at the point $p_i$ and located in the $(r_{m} , \theta_{n})$ bin. For instance, $H_{SC} (p_i , r_{m}, \theta_{n}) = 10 $ means that there are ten other sample points around the point $p_i$ of the skeleton, in the distance range of $r_{m-1} \leq r < r_m$ and in the angle range of $\theta_{n-1} \leq \theta < \theta_{n}$, where 
$m$ is an integer within $[1, 4]$ which means the radius of a circle and $n$ is an integer within $[1, 12]$ which equally divides a circle into 12 sectors, as shown in Fig.~\ref{fig:skeleton_context}. That is, the skeleton context is calculated as
\begin{equation}
\begin{aligned}
\begin{array}{l}
H_{SC}(p_{i}, r_{m}, \theta_{n})=|\operatorname{Bin}(p_{i}, r_{m}, \theta_{k_{n}})|, \\
\operatorname{Bin}(p_{i}, r_m, \theta_n)=\{q \in \mathbb{S} \mid (r_{m-1} \leq \|q-p_{i} \|_{2}<r_{m}) \\
\cap (\theta_{n-1} \leq \angle(q, p_{i})<\theta_{n})\}, \\
\end{array}
\end{aligned}
\label{eq:skeleton_context}
\end{equation}
where $|\cdot|$ shows the number of members in a set, $\mathbb{S}$ is the set of sample points on the skeleton, and $\angle\left(q, p_{i}\right)$ calculates the angle of a vector from $p_i$ to $q$ with respect to the horizontal coordinate. This log-polar histogram located at each point measures the distribution of other points on the skeleton with respect to the center point. By applying this calculation to all sample points of a skeleton for each class $k$, we obtain the skeleton descriptor $\mathcal{SC}^{(k)}$ for the next matching step.

%Therefore, it could play the object descriptor role for the shape matching purposes. As a result, we could use these descriptors as the feature data for matching cost in the next step. The skeleton context can be calculated using the following equation:

\noindent
\textbf{Prototype extraction.} To match with the provided segmentation masks, we first summarize these masks with several shape representatives, that is, extracting prototypes for each class, as shown in Fig.~\ref{fig:prototypes}. We employ the K-medoids algorithm~\cite{rdusseeun1987clustering} to find $K_p$ prototypes for matching. For each class, after initialization with $K_p$ initial medoids, the K-medoids algorithm iterates between the following two steps:
\begin{itemize}[noitemsep,nolistsep]
    \item \textbf{Assignment}: By treating the skeleton descriptor of each shape's $k$-th class $\mathcal{SC}^{(k)}$ as a whole, we assign each skeleton descriptor to its closest medoid $\bar{m}_{i}$, if and only if the distance between them satisfies:
    \begin{equation}
    \begin{aligned}
    D(\mathcal{SC}^{(k)}, \bar{m}_{i}) \leq D(\mathcal{SC}^{(k)}, \bar{m}_{j}), \forall j \neq i ,
    \end{aligned}
    \end{equation}
    where $i$ and $j$ are within $[1, K_p]$, and $D(\cdot,\cdot)$ is a distance metric between two vectors, which is the matching cost computed based on Eq.~\ref{eq:matching_cost}.
    %, $\mathcal{SC}$ is the descriptor vector of a shape, and $\bar{m}_{j}$ is a medoid.
    \item \textbf{Update}: After assigning each skeleton descriptor to a medoid, we have $K_p$ updated clusters. We update the medoid of each cluster by estimating a new descriptor that has the minimum sum of distances to all other skeleton descriptors in its cluster. 
\end{itemize}

\noindent
\textbf{Matching Cost.}
After having the skeleton context $\mathcal{SC}^{(k)}$ for each class $k$ of the predicted segmentation mask and its corresponding $K_p$ skeleton prototypes. Next, we find the closest prototype for the skeleton context of each predicted mask and follow~\cite{belongie2002shape} to measure how close they are. Assume $p_i^1$ and $p_j^2$ are two points from these two skeletons, respectively, based on Eq.~\ref{eq:skeleton_context} we use the following normalized difference to measure their similarity between points on the pair of skeleton contexts:
\begin{equation} 
\begin{aligned}      
C\left(p_{i}^{1}, p_{j}^{2}\right)=\frac{1}{2} \sum_{m, n} \frac{\left(H_{S C}\left(p_{i}^{1}, r_{m}, \theta_{n}\right)-H_{S C}\left(p_{j}^{2}, r_{m}, \theta_{n}\right)\right)^{2}}{H_{S C}\left(p_{i}^{1}, r_{m}, \theta_{n}\right)+H_{S C}\left(p_{j}^{2}, r_{m}, \theta_{n}\right)} .
\end{aligned}
\label{eq:matching_cost}
\end{equation}
By summing up the difference of all sample points on two skeletons, we obtain the matching cost between $\mathcal{SC}^{(k)}$ and its closest one among $K_p$ prototypes $\{\mathcal{SC}_{z}^{(k)}\}_{z = 1}^{K_p}$, that is,
\begin{equation}
    MC(\mathcal{SC}^{(k)}, \{\mathcal{SC}_{z}^{(k)}\}_{z = 1}^{K_p}) = \min_z\sum_{(p_i, p_j)}C(\mathcal{SC}^{(k)}(p_i), \mathcal{SC}_{z}^{(k)}(p_j)).
\end{equation}

\subsubsection{Regularization With Shape Priors} 
We use the above two shape descriptors based on shape contexts, i.e., the class ratio $\mathcal{R}$ and the average distance to the centroid $\mathcal{D}$, and one shape descriptor based on skeleton context $\mathcal{SC}$, to incorporate the shape prior information collected from the provided segmentation masks from a different dataset. 

Given a prediction $s_{\boldsymbol{\theta}}$, we estimate its shape descriptor $\hat{\mathcal{R}}(s_{\boldsymbol{\theta}})$ and $\hat{\mathcal{D}}(s_{\boldsymbol{\theta}})$, and compare them with those given unpaired shapes. In particular, we use a KL divergence to measure the class ratio distribution:
\begin{equation}
\begin{aligned}
\mathcal{L}_{shape}(s_{\boldsymbol{\theta}})=\sum_{k=1}^{K} \operatorname{KL}\left(\hat{\mathcal{R}}^{(k)}(s_{\boldsymbol{\theta}}), {\mathcal{R}}^{(k)}\right).
\end{aligned}
\end{equation}
Here, the class indicator $k \in 1, 2, \cdots, K$.
To reduce the computation cost, i.e., reducing the number of shapes involved in computing $\mathcal{R}^{(k)}$, we only consider those shapes that have a similar averaged distance $\mathcal{D}$ to the predicted one $s_{\boldsymbol{\theta}}$, e.g., their $\mathcal{D}$ difference is less than 0.1:
\begin{equation}
\begin{aligned}
 \min_{\theta}\quad & \mathcal{L}_{shape}(s_{\boldsymbol{\theta}}) \\
                & s.t. \sum_{k=1}^{K} \left|\hat{\mathcal{D}}^{(k)}(s_{\boldsymbol{\theta}})-{\mathcal{D}}^{(k)}\right| \leq 0.1 .
\end{aligned}
\label{eq:lsp_d}
\end{equation}

This minimization is typically handled by using the Lagrangian dual, which is relaxed to an unconstrained optimization via a soft penalty. That is, we integrate the distance constraint via a quadratic penalty, resulting in the unconstrained objective below: 
\begin{equation}
\begin{aligned}
\mathcal{L}_{shape}(s_{\boldsymbol{\theta}}) & =  \sum_{k} \operatorname{KL}\left(\hat{\mathcal{R}}^{(k)}(s_{\boldsymbol{\theta}}), {\mathcal{R}}^{(k)}\right) \\
                          & + \lambda \sum_{k} \mathcal{F}(\hat{\mathcal{D}}^{(k)}(s_{\boldsymbol{\theta}}), {\mathcal{D}}^{(k)}).
\end{aligned}
\end{equation}
Here, $\lambda$ is a weight hyper-parameter to balance these two terms and $\mathcal{F}$ is the quadratic penalty function, i.e., $\mathcal{F}(m_1,m_2)=[m_1-0.9m_2]^2+[1.1m_2-m_1]^2$.
% and \fixme{$[m]_{+}=max(0,m)$}. 

Next, we consider the skeleton descriptor and use the skeleton matching cost as a regularizer:
\begin{equation}
\begin{aligned}
\mathcal{L}_{skeleton}(s_{\boldsymbol{\theta}})= \sum_k MC(SC(s_{\boldsymbol{\theta}}), \{\mathcal{SC}_z\}_z^{K_p \times K}),
\end{aligned}
\end{equation}
where $K_p$ is the number of prototypes and $K$ is the number of segmentation classes. 

Hence, the shape prior loss is defined as:
\begin{equation}
\begin{aligned}
\mathcal{L}_{SP}= L_{shape} + L_{skeleton}
\end{aligned}
\end{equation}
By collecting all the loss terms, we have the final objective function as follows: 
\begin{equation}
\begin{aligned}
    \mathcal{L}_{total} & =  \mathcal{L}_{seg}(M^{init}, M^{pseudo}, M^{voxel}) \\ 
    & + \mathcal{L}_{seg}(M^{final}, M^{pseudo}, M^{voxel}) \\
    & + \beta_1 \mathcal{L}_{bry}(b, B) + \beta_2 \mathcal{L}_{AB} + \beta_3 \mathcal{L}_{SP}.
\end{aligned}
\label{eq:final_loss}
\end{equation}
Here, $\beta_1$, $\beta_2$, and $\beta_3$ are weights for balancing these terms, and their default value is set as 0.3.

\section{Experiments}
\subsection{Datasets and Experimental Settings}

\textbf{ACDC Dataset~\cite{bernard2018deep}.}
This dataset consists of Cine MR images collected from 100 patients by using various 1.5T and 3T MR scanners and different temporal resolutions. For each patient, manual annotations of the right ventricle (RV), left ventricle (LV) and myocardium (MYO) are provided for both the end-diastolic (ED) and end-systolic (ES) phase. The slice size is $256 \times 208$ with the pixel spacing varying from 1.37 to 1.68$mm$. The number of slices is between 28 and 40, and the slice thickness is 5$mm$ or 8$mm$. Following~\cite{valvano2021learning}, we {\it subject-wisely} divide the ACDC dataset into sets of 70\%, 15\% and 15\% for training, validation, and test, respectively. To compare with the previous state-of-the-art methods, which use unpaired masks to learn shape priors, we further divided the training set into two halves, i.e., 35 training images with scribble labels and 35 mask images with segmentation labels. 

\textbf{VS Dataset~\cite{shapey2021segmentation}.}
This dataset collects T2-weighted MRIs from 242 patients with a single sporadic vestibular schwannoma (VS) tumor.
%were acquired in axial view before radiosurgery treatment, 
The size of an image slice is $384 \times 384$ or $448 \times 448$, with a pixel spacing of $~0.5 \times 0.5mm^2$. The number of slices varies from 19 to 118, with a thickness of $1.5mm$. The VS tumor masks are manually annotated by neurosurgeons and physicists. The dataset is {\it subject-wisely} split into 172 for training, 20 for validation, and 46 for testing.

\textbf{CHAOS Dataset~\cite{kavur2021chaos}.}
This dataset has abdominal T1-weighted MR images collected from 20 subjects and the corresponding segmentation masks for liver, kidneys, and spleen. The image slice size is $256 \times 256$ with a resolution of  $1.36-1.89mm$ (average $1.61mm$). The number of slices is between 26 and 50 (average 36) with the slice thickness varying from 5.5 to 9$mm$ (average 7.84 $mm$). We also {\it subject-wisely} divide this dataset into sets of 70\%, 15\%, and 15\% for training, validation, and testing, respectively.

\textbf{Pituitary Microadenoma Dataset.}
To test the performance of our algorithm in practice, we evaluate it on a dataset collected from Ruijin Hospital, Shanghai for the segmentation task of the pituitary with microadenoma lesions. This dataset includes 256 T1-weighted augmented MRIs collected from 86 patients with pituitary microadenoma, consisting of a sequence of coronal slices of brains. The dimension of each image slice varies, including $448\times 448$, $512\times 512$, $768\times 768$, $384\times 384$, $360\times 360$, $256 \times 228$, $336\times 336$ or $256\times 256$, with the pixel spacing ranging from 0.19 to 0.70mm. The number of slices varies from 5 to 16, and the slice thickness is 3mm or 1mm. The dataset is {\it subject-wisely} split into 236 for training with scribble annotations and 20 for testing
with binary masks of the pituitary with lesions. The scribble annotations and segmentation masks are provided by experts. 

 %And the scribble annotations for training and the binary segmentation masks of the pituitary and lesions for testing are provided by experts.

\textbf{M\&Ms Dataset.} To provide shape prior for cardiac segmentation on the ACDC dataset, we choose M\&Ms as a source for learning shape knowledge about ROIs. This dataset is composed of 375 patients with hypertrophic and dilated cardiomyopathies, as well as healthy subjects. All subjects were scanned in clinical centers in three different countries (Spain, Germany, and Canada) using four different magnetic resonance scanner vendors (Siemens, General Electric, Philips, and Canon). The slice size is $256 \times 216$ with the pixel spacing varying from 1.20 to 1.46$mm$.

\newsavebox\CBox
\def\textBF#1{\sbox\CBox{#1}\resizebox{\wd\CBox}{\ht\CBox}{\textbf{#1}}}

\begin{table*}[t]
\centering
\begin{center}
\scriptsize
\renewcommand\arraystretch{1.2} 
\caption{Quantitative comparison among baselines and our method for volumetric segmentation on three datasets. Mean and standard deviation (subscript) are reported. The upper bounds are colored in {\color{blue}blue}, and the best results by using scribbles are marked in \textbf{bold}. $^\dagger$P is short for Point, indicating the extreme points. We have such annotations only for the VS dataset. $^*$These numbers are taken from the InExtremeIS paper. (Best viewed in color)}
\begin{tabular}{p{0.2cm}|p{0.2cm}|p{2.1cm}|p{1.1cm}p{1.1cm}p{1.2cm}|p{1.1cm}p{1.1cm}p{1.2cm}|p{1.1cm}p{1.1cm}p{1.2cm}}
% \toprule
% \multicolumn{3}{c}{} &\multicolumn{9}{c}{Dataset} \\
% \cmidrule[0.2pt]{4-12}
\toprule
\multicolumn{3}{c|}{\multirow{3}*{\diagbox{Approach}{Dataset}}} &\multicolumn{3}{c|}{ACDC} & \multicolumn{3}{c|}{VS} & \multicolumn{3}{c}{CHAOS} \\ 
\cmidrule{4-12}
\multicolumn{3}{c|}{} & \makecell{Dice $\uparrow$ \\ (\%)} & \makecell{HD95 $\downarrow$ \\ (mm)} & \makecell{Precision $\uparrow$ \\(\%)} & \makecell{Dice $\uparrow$ \\ (\%)} & \makecell{HD95 $\downarrow$ \\ (mm)} & \makecell{Precision $\uparrow$\\(\%)} & \makecell{Dice $\uparrow$\\ (\%)} & \makecell{HD95 $\downarrow$ \\ (mm)} & \makecell{Precision $\uparrow$ \\(\%)} \\
\cmidrule{1-12}
\multirow{13}*{\begin{sideways} Supervision Type \end{sideways}} & \multirow{8}*{\begin{sideways} Scribble \end{sideways}} & $\rm UNet_{PCE}$~\cite{tang2018normalized} & \makecell[c]{$79.0_{\pm06}$} & \makecell[c]{$6.9_{\pm04}$} & \makecell[c]{$77.3_{\pm06}$} & \makecell[c]{$44.6_{\pm08}$} & \makecell[c]{$6.5_{\pm03}$} & \makecell[c]{$43.8_{\pm05}$} & \makecell[c]{$34.4_{\pm06}$} & \makecell[c]{$9.4_{\pm03}$} & \makecell[c]{$36.6_{\pm05}$}  \\
& & ConstrainedCNN~\cite{kervadec2019constrained}& \makecell[c]{$80.1_{\pm04}$} & \makecell[c]{$5.4_{\pm05}$} & \makecell[c]{$79.8_{\pm05}$} & \makecell[c]{$68.1_{\pm04}$} & \makecell[c]{$7.1_{\pm04}$} & \makecell[c]{$67.7_{\pm04}$} & \makecell[c]{$62.1_{\pm04}$} & \makecell[c]{$6.6_{\pm04}$} & \makecell[c]{$65.1_{\pm04}$}  \\
& & MAAG~\cite{valvano2021learning}& \makecell[c]{$83.4_{\pm04}$} & \makecell[c]{$8.6_{\pm04}$} & \makecell[c]{$78.5_{\pm05}$} & \makecell[c]{$69.4_{\pm06}$} & \makecell[c]{$5.9_{\pm05}$} & \makecell[c]{$56.8_{\pm05}$} & \makecell[c]{$66.4_{\pm05}$} & \makecell[c]{$3.8_{\pm05}$} & \makecell[c]{$57.2_{\pm06}$}  \\
& & ScribbleSeg~\cite{luo2022scribble}& \makecell[c]{$87.2_{\pm07}$} & \makecell[c]{$9.3_{\pm05}$} & \makecell[c]{$86.8_{\pm05}$} & \makecell[c]{$80.6_{\pm04}$} & \makecell[c]{$8.2_{\pm04}$} & \makecell[c]{$79.0_{\pm04}$} & \makecell[c]{$77.1_{\pm04}$} & \makecell[c]{$4.1_{\pm04}$} & \makecell[c]{$72.3_{\pm04}$}  \\
\cmidrule{3-12}
& & Ours w/o PLPM           & \makecell[c]{$83.2_{\pm05}$} & \makecell[c]{$7.7_{\pm03}$} & \makecell[c]{$84.1_{\pm05}$} & \makecell[c]{$78.8_{\pm05}$} & \makecell[c]{$\boldsymbol{4.6_{\pm01}}$} & \makecell[c]{$77.6_{\pm05}$} & \makecell[c]{$81.2_{\pm07}$} & \makecell[c]{$5.8_{\pm08}$} & \makecell[c]{$82.0_{\pm06}$}  \\
& & Ours w/o SBPM          & \makecell[c]{$85.6_{\pm05}$} & \makecell[c]{$4.6_{\pm04}$} & \makecell[c]{$85.5_{\pm04}$} & \makecell[c]{$80.6_{\pm05}$} & \makecell[c]{$7.1_{\pm03}$} & \makecell[c]{{$\boldsymbol{81.6_{\pm04}}$}} & \makecell[c]{$84.6_{\pm05}$} & \makecell[c]{$5.5_{\pm05}$} & \makecell[c]{$83.1_{\pm05}$}  \\
& & Ours w/o ABL         & \makecell[c]{$88.7_{\pm04}$} & \makecell[c]{$5.1_{\pm08}$} & \makecell[c]{\textBF{$86.0_{\pm05}$}} & \makecell[c]{$81.0_{\pm03}$} & \makecell[c]{$4.8_{\pm01}$} & \makecell[c]{$80.1_{\pm05}$} & \makecell[c]{$85.6_{\pm04}$} & \makecell[c]{$4.8_{\pm05}$} & \makecell[c]{$81.3_{\pm02}$}  \\
& & Scribble2D5 (ours)    & \makecell[c]{\textBF{$90.6_{\pm03}$}} & \makecell[c]{\textBF{$2.3_{\pm05}$}} & \makecell[c]{$84.7_{\pm05}$} & \makecell[c]{$\boldsymbol{82.6_{\pm07}}$} & \makecell[c]{$4.7_{\pm04}$} & \makecell[c]{$81.5_{\pm06}$} & \makecell[c]{{$\boldsymbol{86.0_{\pm04}}$}} & \makecell[c]{{$\boldsymbol{2.9_{\pm02}}$}} & \makecell[c]{$\boldsymbol{88.2_{\pm03}}$}  \\
& & Scribble2D5 w/ SP    & \makecell[c]{$\boldsymbol{92.2_{\pm04}}$} & \makecell[c]{{$\boldsymbol{1.1_{\pm01}}$}} & \makecell[c]{$\boldsymbol{88.6_{\pm05}}$} & \makecell[c]{--} & \makecell[c]{--} & \makecell[c]{--} & \makecell[c]{--} & \makecell[c]{--} & \makecell[c]{--}  \\
\cmidrule{2-12}
& \begin{sideways}$\text{P}^{\dagger}$\end{sideways} & InExtremeIS~\cite{dorent2021inter}   & \makecell[c]{--} & \makecell[c]{--} & \makecell[c]{--} & \makecell[c]{$81.9_{\pm03}^{*}$} & \makecell[c]{\color{blue}{{$3.7_{\pm03}^{*}$}}} & \makecell[c]{\color{blue}{$92.9_{\pm02}^{*}$}}  & \makecell[c]{--} & \makecell[c]{--} & \makecell[c]{--} \\ 
\cmidrule{2-12}
& \multirow{2}*{\begin{sideways} Mask \end{sideways}} & 2D UNet~\cite{ronneberger2015u} & \makecell[c]{$93.0_{\pm05}$} & \makecell[c]{$3.5_{\pm15}$} & \makecell[c]{$90.2_{\pm07}$} & \makecell[c]{$80.4_{\pm03}$} & \makecell[c]{$7.3_{\pm04}$} & \makecell[c]{$81.2_{\pm03}$} & \makecell[c]{$82.3_{\pm04}$} & \makecell[c]{$3.3_{\pm01}$} & \makecell[c]{$81.7_{\pm05}$} \\
% & & 3D U-Net\cite{shapey2019artificial}  & \makecell[c]{-} & \makecell[c]{-} & \makecell[c]{-} & \makecell[c]{$83.6_{03}$} & \makecell[c]{-} & \makecell[c]{-} & \makecell[c]{-} & \makecell[c]{-} & \makecell[c]{-}\\
& & 2.5D UNet~\cite{shapey2019artificial} & \makecell[c]{\color{blue}{$96.1_{\pm03}$}} & \makecell[c]{\color{blue}{$0.3_{\pm00}$}} & \makecell[c]{\color{blue}{$95.3_{\pm04}$}} & \makecell[c]{\color{blue}{$87.3_{\pm02}$}} & \makecell[c]{$6.8_{\pm04}$} & \makecell[c]{$84.7_{\pm03}$} & \makecell[c]{\color{blue}{$90.8_{\pm03}$}} & \makecell[c]{\color{blue}{$1.1_{\pm00}$}} & \makecell[c]{\color{blue}{$91.4_{\pm05}$}} \\
% \cmidrule[0.8pt]{2-12}
\bottomrule
\end{tabular}
\label{tab:results}
\end{center}
\end{table*}

\begin{table}[t]
\centering
\begin{center}
% \scriptsize
\renewcommand\arraystretch{1.2} 

\caption{Quantitative comparison among baselines and our method for the pituitary and microadenoma segmentation on our private dataset. Mean and standard deviation (subscript) are reported, and the best results are in \textbf{bold}.}
\begin{tabular}{p{2.4cm}|p{1.2cm}p{1.2cm}p{1.3cm}}
\toprule
 %&\multicolumn{3}{c}{Pituitary and Microadenoma} \\ 
%\cmidrule{2-4}
\makecell[l]{Method} & \makecell{Dice $\uparrow$ \\ (\%)} & \makecell{HD95 $\downarrow$ \\ (mm)} & \makecell{Precision $\uparrow$ \\(\%)}  \\
\cmidrule{1-4}
$\rm UNet_{PCE}$~\cite{tang2018normalized} & \makecell[c]{$63.0_{\pm06}$} & \makecell[c]{$6.9_{\pm04}$} & \makecell[c]{$67.3_{\pm06}$}   \\
%ConstrainedCNN~\cite{kervadec2019constrained}& \makecell[c]{?} & \makecell[c]{?} & \makecell[c]{?}   \\
MAAG~\cite{valvano2021learning}& \makecell[c]{$75.6_{\pm04}$} & \makecell[c]{$7.6_{\pm04}$} & \makecell[c]{$74.5_{\pm05}$}   \\
\hline
Ours w/o LPM         & \makecell[c]{$72.1_{\pm05}$} & \makecell[c]{$5.5_{\pm03}$} & \makecell[c]{$74.1_{\pm05}$}   \\
Ours w/o SBPM          & \makecell[c]{$74.6_{\pm05}$} & \makecell[c]{$3.8_{\pm04}$} & \makecell[c]{$75.8_{\pm04}$}   \\
Ours w/o ABL         & \makecell[c]{$76.7_{\pm04}$} & \makecell[c]{$5.1_{\pm08}$} & \makecell[c]{\textBF{$76.0_{\pm05}$}}  \\
Scribble2D5 (ours)     & \makecell[c]{{$\boldsymbol{78.8_{\pm03}}$}} & \makecell[c]{{$\boldsymbol{2.3_{\pm05}}$}} & \makecell[c]{$\boldsymbol{77.7_{\pm05}}$} \\
\bottomrule
\end{tabular}
\label{tab:private_results}
\end{center}
\end{table}

\textbf{Scribble Generation.} For the ACDC dataset, we use the scribbles provided in~\cite{valvano2021learning}, which are manually drawn by experts at both end-diastolic and end-systolic phases. For both VS and CHAOS datasets, following~\cite{rajchl2017employing}, we simulate scribbles by an iterative morphological erosion and closing of segmentation masks, which results in a one-pixel skeleton for each object. Since the resulting background scribble is winding, we use the ITK-SNAP tool to annotate the background with 1-pixel width curves.

%For Pituitary Microadenoma dataset, researchers first draw the scribbles needed by learning from the doctors and the masks provided previously. 

\textbf{Training Details.} For all public datasets, we randomly crop an image volume and obtain patches of size $224 \times 224 \times 32$ as the network inputs for training. For our private dataset, the patch size is $192 \times 192 \times 8$. If an input image has a smaller size in one or more dimensions, we pad it with zeros to match the input size. 
At the inference stage, we use a sliding window when an image has a larger input size than intputs, with 25\% of patch size overlaps at the borders.

For all public datasets, we train our model for 200 epochs with early stopping. 
The weights of the network are initialized by following a normal distribution with a mean of 0 and a variance of 0.01. We use Adam optimizer with a weight decay $10^{-7}$ and an initial learning rate 1$e$-4. The whole training takes about 6 hours with a batch size of 4 on one NVIDIA GeForce RTX 3090 GPU. 
Differently, for our private dataset, we train the models for 50 epochs with early stopping and an Adam optimizer with a weight decay 2$e$-7. Since the pituitary tumors have irregular shapes, while the active boundary loss smoothes out the predicted boundary, we set its weight $\beta_2$ as 0. 

%. The weights of the network are initialized by following a normal distribution with a mean of 0 and variance of 0.01. We use Adam optimizer with a weight decay 2e-7 and an initial learning rate 1e-4. Although the active boundary loss benefits the other three databases as it considers the shape and intensity of 3D images, it is not applicable for the Pituitary Microadenoma dataset since the targets of patients with pituitary tumors often have irregular shapes. And the shapes will be smoothed if the active boundary loss is used. For this reason, AB loss is eliminated for this dataset. The whole training takes about 3 hours with a batch size of 8 on one NVIDIA GeForce RTX 3090 GPU. 
%Table x presents our experimental results on this dataset. 

\begin{table}[t]
\centering
\caption{Performance comparison in Dice score (\%) on the ACDC dataset between our Scribble2D5 and current weakly-supervised methods. We borrow the segmentation results reported in~\cite{zhang2022cyclemix} for comparison.}
 % \fixme{Does CycleMix use shape prior?}
\begin{tabular}{p{2.2cm}|p{1.8cm}|p{0.4cm}<{\centering}p{0.4cm}<{\centering}p{0.4cm}<{\centering}|p{0.4cm}<{\centering}}
\toprule
Method & Data & LV & MYO & RV & Avg. \\
\cmidrule{1-6}
\multicolumn{6}{l}{35 scribbles} \\
\cmidrule{1-6}
${\rm UNet_{PCE}}$~\cite{tang2018normalized} & scribbles & 84.2 & 76.4 & 69.3 & 76.6 \\
${\rm UNet_{WPCE}}$~\cite{valvano2021learning} & scribbles & 78.4 & 67.5 & 56.3 & 67.4 \\
${\rm UNet_{CRF}}$~\cite{zheng2015conditional} & scribbles & 76.6 & 66.1 & 59.0 & 67.2 \\
CycleMix~\cite{zhang2022cyclemix} & scribbles & 88.3 & 79.8 & 86.3 & 84.8 \\
Scribble2D5 (ours) & scribbles & \textbf{92.3} & \textbf{82.2} & \textbf{89.8} & \textbf{88.1} \\
\cmidrule{1-6}
\multicolumn{6}{l}{35 scribbles + 35 unpaired masks} \\
\cmidrule{1-6}
${\rm UNet_{D}}$~\cite{valvano2021learning} & scribbles+masks & 40.4 & 59.7 & 75.3 & 58.5 \\
PostDAE~\cite{larrazabal2020post} & scribbles+masks & 80.6 & 66.7 & 55.6 & 67.6 \\
ACCL~\cite{zhang2020accl} & scribbles+masks & 87.8 & 79.7 & 73.5 & 80.3 \\
MAAG~\cite{valvano2021learning} & scribbles+masks & 87.9 & 81.7 & 75.2 & 81.6 \\
ours w/ SP & scribbles+masks & \textbf{94.2} & \textbf{84.1} & \textbf{92.0} & \textbf{90.1} \\
\bottomrule
\end{tabular}
\label{tab:shape_prior}
\end{table}

\begin{figure*}[t]
\begin{tabular}
{p{1.4cm}<{\centering}p{1.4cm}<{\centering}p{1.4cm}<{\centering}p{1.4cm}<{\centering}p{1.4cm}<{\centering}p{1.4cm}<{\centering}p{1.4cm}<{\centering}p{1.4cm}<{\centering}p{1.4cm}<{\centering}p{1.4cm}<{\centering}}
\footnotesize{Image} & \footnotesize{Scribbles} & \footnotesize{GT} & \footnotesize{$\text{UNet}_\text{PCE}$} & \footnotesize{MAAG} & \footnotesize{ScribbleSeg} & \footnotesize{CycleMix} & \footnotesize{2D UNet} & \footnotesize{2.5D UNet} & \footnotesize{Ours} \\
\end{tabular}
\includegraphics[width=\textwidth]{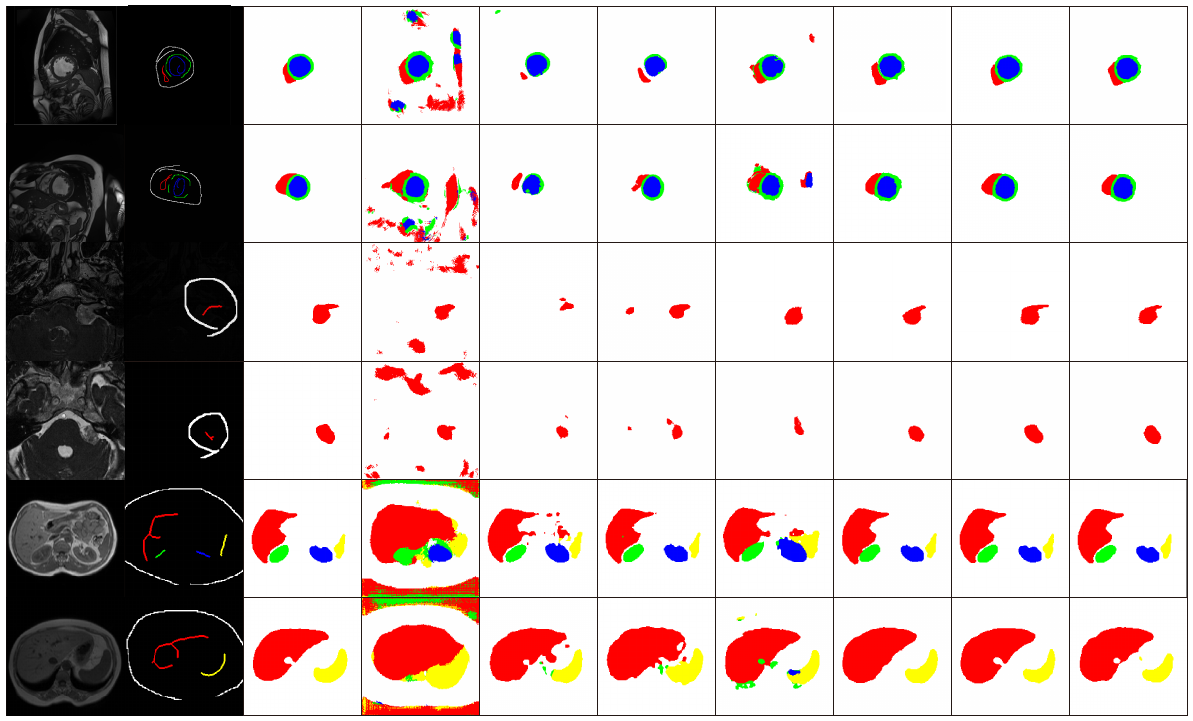}
\caption{Qualitative comparison among baseline methods and ours on the ACDC, VS, and CHAOS datasets. 
\textbf{ACDC}: \textcolor{red}{Red}: LV, \textcolor{green}{green}: MYO, \textcolor{blue}{blue}: RV;
\textbf{VS}: \textcolor{red}{Red}: vestibular schwannoma tumor;
\textbf{CHAOS}: \textcolor{red}{Red}: Liver, \textcolor{green}{green}: left kidney, \textcolor{blue}{blue}: right kidney, \textcolor{yellow}{yellow}: spleen; for all, white indicates the background. (Best viewed in color) 
}

%(a) Image. (b) Ground truth. (c) Scribble annotation. (d) $\text{UNet}_\text{PCE}$. (e) MAAG. (f) ScribbleSeg. (g) CycleMix. (h) 2D-UNet. (i) 2.5D-UNet. (j) Scribble2D5(ours).} 
\label{fig:qualitive}
\end{figure*}

\textbf{Baselines and Evaluation Metrics.}
To demonstrate the effectiveness of our methods, we select three groups of baselines, including two fully-supervised methods (i.e., 2D UNet~\cite{ronneberger2015u} and 2.5D UNet~\cite{shapey2019artificial}), nine weakly-supervised methods using scribbles (i.e., $\text{UNet}_\text{PCE}$~\cite{tang2018normalized}, ${\text{UNet}_\text{WPCE}}$~\cite{valvano2021learning}, 
${\text{UNet}_\text{CRF}}$~\cite{zheng2015conditional},
${\text{UNet}_\text{D}}$~\cite{valvano2021learning}, 
MAAG~\cite{valvano2021learning}, ScribbleSeg~\cite{luo2022scribble}, CycleMix~\cite{zhang2022cyclemix}, PostDAE~\cite{larrazabal2020post}, and ACCL~\cite{zhang2020accl}) and one weakly-supervised method using extreme points~\cite{dorent2021inter}. To evaluate the segmentation performance, we use three metrics, i.e., the Dice score to calculate the overlap between our prediction and the ground truth (GT) segmentation mask, the 95th percentile of the Hausdorff Distance (HD95) to measure the distance between our boundary and GT's, and the precision to check the purity of the positively-segmented voxels.

% \begin{figure*}[t]
% \includegraphics[width=\textwidth]{figures/VS-qualitative-comparison.pdf}
% \caption{Qualitative results using different methods on the VS dataset. (a) Image. (b) Ground truth. (c) Scribble annotation. (d) $\text{UNet}_\text{PCE}$. (e) MAAG. (f) ScribbleSeg. (g) CycleMix. (h) 2D-UNet. (i) 2.5D-UNet. (j) Scribble2D5(ours).} 
% \label{fig:qualitive-vs}
% \end{figure*}

% \begin{figure*}[t]
% \includegraphics[width=\textwidth]{figures/CHAOS-qualitative-comparison.pdf}
% \caption{Qualitative results using different methods on the CHAOS dataset. (a) Image. (b) Ground truth. (c) Scribble annotation. (d) $\text{UNet}_\text{PCE}$. (e) MAAG. (f) ScribbleSeg. (g) CycleMix. (h) 2D-UNet. (i) 2.5D-UNet. (j) Scribble2D5(ours).} 
% \label{fig:qualitive-chaos}
% \end{figure*}

\begin{figure*}[t]
\centering
\begin{tabular}
{p{1.0cm}<{\raggedleft}p{2.0cm}<{\raggedleft}p{1.6cm}<{\raggedleft}p{2.4cm}<{\raggedleft}p{1.8cm}<{\raggedleft}p{1.8cm}<{\raggedleft}}
\footnotesize{Image} & \footnotesize{Scribble} & \footnotesize{GT} & \footnotesize{$\text{UNet}_\text{PCE}$} & \footnotesize{MAAG} & \footnotesize{Ours} \\
\end{tabular}
\includegraphics[width=0.8\textwidth]{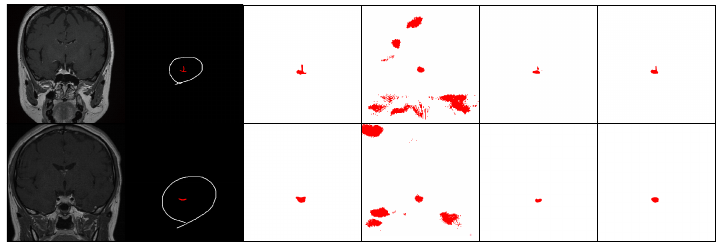}
\caption{Qualitative comparison between our Scribble2D5 and two baselines on our private Pituitary Microadenoma dataset.
% \fixme{Is there any difference in the results? Keep consistent with others, scribble goes first. We only need 2-3 samples to show the difference.}
}
\label{fig:qualitive-ruijin}
\end{figure*}

\begin{figure*}[t]
\includegraphics[width=\textwidth]{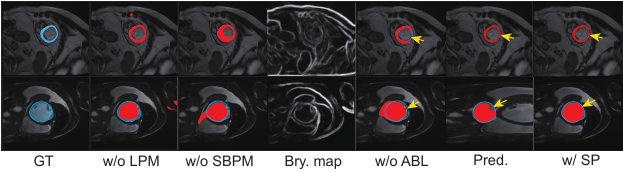}
\caption{Visualization of Scribble2D5's intermediate and final results on images sampled from the ACDC dataset. The ground truth (GT) is colored in blue, like the blue region in the first column and the blue contours overlaid on other images, and our predictions are colored in red. The yellow arrows show the effect of the active boundary loss (ABL) and considering shape prior (SP). (Best viewed in color)}
\label{fig:ablation}
\end{figure*}

\subsection{Experimental Results}

\subsubsection{Comparison with SOTA methods}
Table~\ref{tab:results},~\ref{tab:private_results}, and~\ref{tab:shape_prior} present our experimental results on three public datasets and one private dataset with a comparison to our baselines. 

For ACDC, VS, and CHAOS datasets, the upper bounds of the segmentation performance are mainly provided by the 2.5D UNet, which are colored in blue in Table~\ref{tab:results}. Compared to the scribble-based SOTA method on ACDC and CHAOS datasets, i.e., ScribbleSeg~\cite{luo2022scribble}, scribble2D5 improves the Dice score by $5\%$ and $8.9\%$, reduces the HD95 by $8.2$mm and $1.2$mm, and improves the precision by $1.8\%$ and $15.9\%$, respectively. Compared to the extreme-point-based SOTA method on the VS dataset, i.e., InExtremeIS~\cite{dorent2021inter}, although our method has a lower precision and HD95, it improves the Dice score by $0.7\%$.
We do not report InExtremeIS' results on ACDC and CHAOS datasets because extreme points for these two datasets are not available or easy to generate. 

Figure~\ref{fig:qualitive} visualizes some sample results of our method compared to six baselines. Overall, we have fewer false positives compared to scribble-based methods, i.e., $\text{UNet}_\text{PCE}$ and MAAG, and better boundary localization with more accurate boundary prediction for each ROI. Regarding the comparison with mask-based methods, our method sometimes generates even better masks than 2D UNet, while it still needs improvements in details compared to 2.5D UNet.

For our private dataset, we save the ones with segmentation masks for testing and the ones with scribble annotations for training. Also, we do not have shape prior information about a pituitary with tumors. Therefore, we compare our method with scribble-based methods only. As reported in Table~\ref{tab:private_results}, our method outperforms MAAG, by improving 3.2\% dice score. Figure~\ref{fig:qualitive-ruijin} shows some sample results which demonstrates that our method is better at details.

More comparison results are included in Table~\ref{tab:shape_prior}, which reports the performance comparison on the ACDC dataset between eight baselines and our methods by using 35 images with scribbles and by adding another 35 unpaired masks as shape prior for learning. Our methods (with and without shape priors) outperform baselines by a good margin on both individual segmentation regions and their average.

\subsubsection{Ablation Study}
To check the effectiveness of each module in our method, we perform an ablation study with the following four variants: 

a) \textbf{Ours w/o PLPM}: Scribble2D5 without the pseudo label propagation module (PLPM);

b) \textbf{Ours w/o SBPM}: Scribble2D5 without the static boundary prediction module (SBPM), which removes the static boundary prediction module and active boundary loss;

c) \textbf{Ours w/o ABL}: Scribble2D5 without the active boundary loss (ABL); 

d) \textbf{Scribble2D5 w/ SP}: Scribble2D5 with shape prior (SP) if available. 

The results of the ablation study on both public and private datasets are reported in Table~\ref{tab:results} and Table~\ref{tab:private_results}, respectively.  For the ACDC, CHAOS, and our private dataset, we can observe consistent improvement by adding PLPM, SBPM, and ABL modules, one by one. The ACDC experiment in Table~\ref{tab:shape_prior} also demonstrates the effectiveness of introducing shape prior. Regarding our results on the VS dataset, only the Dice score consistently increases as adding each module gradually; however, the HD95 and precision values are just slightly lower than the highest ones. We still consider our full model performs the best in the ablation study on this dataset. 

%; however, considering all three metrics, the full model has the best performance.

%Take the ACDC dataset for an example, as shown in Table~\ref{tab:results}, without LPM but all others, the Dice score reduces from $90.4\%$ to $80.6\%$. With LPM but without SBPM, the Dice score is $85.6\%$; then the static boundary prediction module contributes an improvement of $3.1\%$, and the active boundary loss contributes an additional improvement of $1.7\%$ in Dice score. 

Figure~\ref{fig:ablation} visualizes two samples from the ACDC dataset with our intermediate and final prediction results. Without PLPM, our method suffers from false positives far away from the ROI; without SBPM, our method has the over-segment issues of the ROI. By adding the boundary map and active boundary regularization, our method adjusts the prediction based on the image edge and texture information. After considering the shape prior, the shape of the ROI is further adjusted towards the true shape, resulting in the closest prediction compared to the ground truth.  

\begin{figure}[t]
\centering
\includegraphics[width=0.48\textwidth]{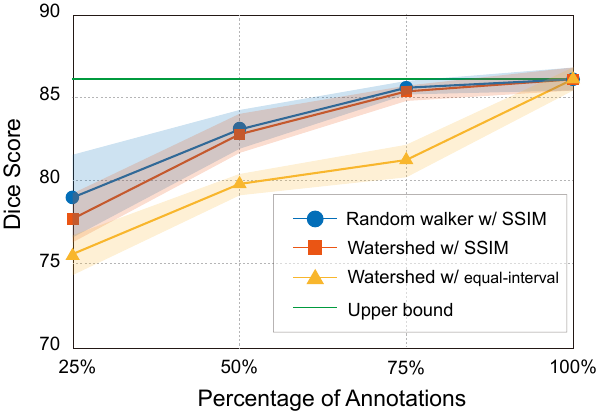}
\caption{Dice score obtained on the label generation and test data by SSIM sampling or equal-interval sampling when changing the percentage of available annotations. We consider 100\% (use all the densely-annotated masks) as the upper bound.}
\label{fig:scrigen}
\end{figure}

\subsubsection{Robustness to Limited Annotations}
% 作为3D scribble，再加一个稀疏程度，只在部分slices上标注，这个slices的比例对结果的影响
Since we work on volumetric image segmentation, each volume has a sequence of 2D slices that need scribble annotations for training. In practice, we probably have missing annotations on some slices. In this experiment, we analyze the robustness of our model with a scarcity of scribble annotations on the ACDC dataset. In this experiment, we only annotate partial 2D Axial slices, e.g., 25\%, 50\%, or 75\% of the image slices of a volume, respectively. 
% These image slices for scribble annotations are sampled from the volume using different techniques, e.g., uniform sampling, and the structural similarity index measure (SSIM) based sampling. 
To generate scribbles on those slices with missing annotations, we explore both watershed and random walker methods. These two methods are based on structural similarity index measure (SSIM) sampling or equal-interval sampling.
%at equal intervals or using SSIM based sampling to screen out the 'annotated group' needed in watershed and random walker, then we perform scribble generation as discussed in Section III.
Table~\ref{tab:different_label_propagation_methods} shows the dice score of our Scribble2D5 using the pseudo labels generated by these two methods with two kinds of sampling strategies. Choosing a good label propagation strategy, like the random walker approach with SSIM sampling, can reduce the annotation amount by 25\% while achieving comparable segmentation accuracy. We do not test our method using the random walker with equal-interval sampling since the watershed experiment shows SSIM is a better sampling choice. 

\begin{table}[t]
\centering
\begin{center}
\scriptsize
\caption{The performance (Dice Scores) on generated scribbles from different label propagation methods.}

\begin{tabular}{lllll}%{p{2.6cm}|p{0.8cm}p{0.8cm}p{0.8cm}|p{0.8cm}}
\toprule
Type of Scribbles & 25\% & 50\% & 75\% & 100\% \\
\cmidrule{1-5}
Random walker w/ SSIM & \textbf{79.5$\pm$3.1} & \textbf{83.7$\pm$2.3} & \textbf{86.0$\pm$1.7} & 86.1$\pm$2.3 \\
Watershed w/ SSIM     & 77.7$\pm$2.3 & \textbf{83.7$\pm$2.1} & 85.8$\pm$1.9 & 86.1$\pm$2.3 \\
Watershed w/ equal-interval & 75.0$\pm$2.4 & 79.1$\pm$1.8 & 81.6$\pm$2.2 & 86.1$\pm$2.3 \\

\bottomrule
\end{tabular}

\vspace{-0.1in}

\label{tab:different_label_propagation_methods}
\end{center}
\end{table}

\begin{figure}[t]
\includegraphics[width=0.49\textwidth]{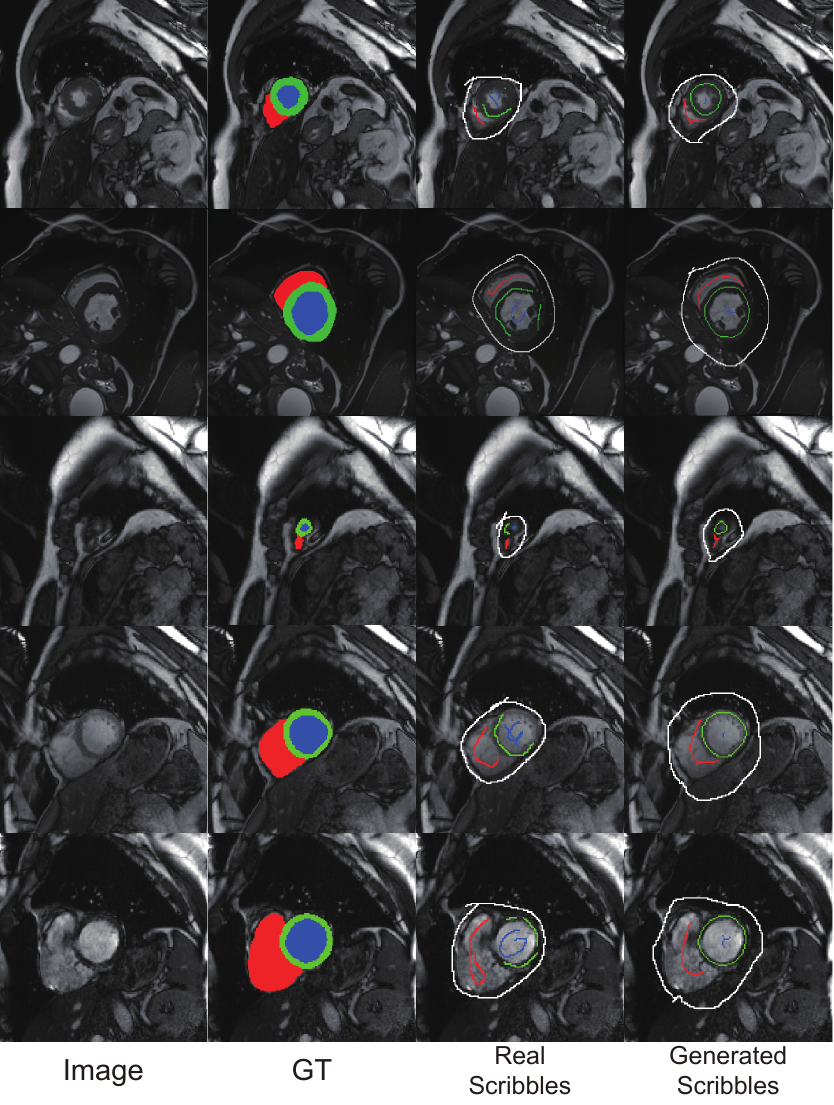}
\caption{Comparison between real and generated scribbles. (Best viewed in color, GT: the ground-truth mask)}
\label{fig:scribble-generation}
\end{figure}

\begin{table}[t]
\centering
\begin{center}
\caption{The performance (Dice Scores) on generated scribbles compared with real scribbles provided by experts.}
\begin{tabular}{lllll}%{p{2.4cm}|p{0.8cm}p{0.8cm}p{0.8cm}|p{0.8cm}}
\toprule
Type of Scribbles & LV & MYO & RV & Avg. \\
\hline
Real      & \textbf{94.3} & \textbf{89.6} & \textbf{88.2} & \textbf{90.7} \\
Generated & 87.9 & 84.2 & 78.4 & 83.5 \\

\bottomrule
\end{tabular}

\vspace{-0.1in}

\label{tab:generated_scribbles_and_real_scribbles_provided_by_experts}
\end{center}
\end{table}

\subsubsection{Comparison between Real and Generated Scribbles}
To further study the possibility of using generated scribbles to replace the real ones, we perform the experiment on the ACDC dataset and compare the manual scribbles annotated by experts and the one generated by simulating scribbles through an iterative morphological erosion and closing of segmentation masks~\cite{rajchl2017employing}. 
Firstly, we measure the size difference between these two scribbles. The manual scribbles annotated for the foreground ROIs occupy 11.7\% of a mask, while the generated ones occupy 7.2\%. That is, the manual scribbles tend to cover more regions of interest. Then, we evaluate the performance difference between them. As shown in Table~\ref{tab:generated_scribbles_and_real_scribbles_provided_by_experts}, using the manual scribbles achieve 90.7\% on average in Dice score, while only 83.5\% by using the generated ones. This is probably because, unlike the manual ones, the generated scribbles locate close to the center lines of ROIs as shown in Fig.~\ref{fig:scribble-generation}, which are far away from the boundary and provide less information about ROIs. Hence, if manual scribbles are available in the VS and CHAOS datasets, the performance of our method has the potential to be further improved. 

%Regarding the performance, using generated scribbles results in a 0.836 dice, while the real ones achieve 0.906 (as shown in Table~\ref{tab:generated_scribbles_and_real_scribbles_provided_by_experts}). The generated scribbles performs worse 

\section{Conclusion and Discussion}
In this paper, we propose a weakly-supervised volumetric image segmentation network, Scribble2D5, which outperforms existing scribble-based methods by a good margin. One limitation of our method is that our pseudo-boundary labels are a stack of pre-computed 2D boundaries, which are not purely 3D and will be explored in the future. We also observe that the shape and location of scribbles would affect the segmentation accuracy, summarizing a couple of rules to make scribble annotations for different ROIs would be useful in practice, which will be left as future work. In addition, there is still a performance gap between our method and fully-supervised segmentation approaches. To further improve the model performance, a possible solution is using interactive segmentation and learning from user feedback, which will be explored in the future. 

\section{Acknowledgements}
This work was supported by NSFC 62203303 and Shanghai Municipal Science and Technology Major Project 2021SHZDZX0102.

\bibliographystyle{IEEEtran}
\bibliography{refs}

\end{document}